\documentclass{article} % For LaTeX2e
\usepackage{iclr2022_conference,times}

\usepackage{hyperref}
\usepackage{url}

\usepackage[utf8]{inputenc} % allow utf-8 input
\usepackage[T1]{fontenc}    % use 8-bit T1 fonts

\usepackage{booktabs}       % professional-quality tables
\usepackage{amsfonts}       % blackboard math symbols
\usepackage{nicefrac}       % compact symbols for 1/2, etc.
\usepackage{microtype}      % microtypography
\usepackage{xcolor}         % colors
\usepackage{graphicx}
\usepackage{amsmath,amssymb,amsfonts}
\usepackage{multirow}
\usepackage{subfig}

\usepackage{algorithm}  
\usepackage{algorithmicx}  
\usepackage{algpseudocode}  

\title{AF$_2$: Adaptive Focus Framework for \\Aerial Imagery Segmentation}

\author{Lin Huang\textsuperscript{\rm 1}, Qiyuan Dong, Lijun Wu\textsuperscript{\rm 1}, Jia Zhang\textsuperscript{\rm 1}, Jiang Bian\textsuperscript{\rm 1}, Tie-Yan Liu\textsuperscript{\rm 1} \\
\textsuperscript{\rm 1}Microsoft Research Aisia, BeiJing, China \\
\texttt{huang.lin@microsoft.com,akaqyd@gmail.com}\\
\texttt{lijun.wu@microsoft.com,jia.zhang@microsoft.com}\\
\texttt{jiang.bian@microsoft.com,tie-yan.liu@microsoft.com} \\
}

\iclrfinalcopy % Uncomment for camera-ready version, but NOT for submission.
\begin{document}

\maketitle

\begin{abstract}
As a specific semantic segmentation task, aerial imagery segmentation has been widely employed in high spatial resolution (HSR) remote sensing images understanding. Besides common issues (e.g. large scale variation) faced by general semantic segmentation tasks, aerial imagery segmentation has some unique challenges, the most critical one among which lies in foreground-background imbalance. There have been some recent efforts that attempt to address this issue by proposing sophisticated neural network architectures, since they can be used to extract informative multi-scale feature representations and increase the discrimination of object boundaries. Nevertheless, many of them merely utilize those multi-scale representations in ad-hoc measures but disregard the fact that the semantic meaning of objects with various sizes could be better identified via receptive fields of diverse ranges. In this paper, we propose Adaptive Focus Framework (AF$_2$), which adopts a hierarchical segmentation procedure and focuses on adaptively utilizing multi-scale representations generated by widely adopted neural network architectures. Particularly, a learnable module, called Adaptive Confidence Mechanism (ACM), is proposed to determine which scale of representation should be used for the segmentation of different objects. Comprehensive experiments show that AF$_2$ has significantly improved the accuracy on three widely used aerial benchmarks, as fast as the mainstream method.

%Besides the large scale variation problems for semantic segmentation in most semantic segmentation datasets, high spatial resolution (HSR) remote sensing imagery segmentation faces its own challenges and the most critical one lies in the foreground-background imbalance. Previous studies mainly focus on design delicate model structure in both general and aerial image segmentation, however, most of them ignore the other side of the coin, that is how to efficiently utilize the multi-scale representation.

%In order to achieve this goal, a learnable module, called Adaptive Confidence Mechanism (ACM), is proposed to determine which scale of representation should be used for the segmentation of different objects.
%Comprehensive experiments show that AF$_2$ has significantly improved the accuracy on three widely used aerial benchmarks, with the fastest convergence speed and inference speed.
\end{abstract}

\section{Introduction}
Understanding geospatial objects, such as plants, buildings, vehicles, etc., in high spatial resolution (HSR) remote sensing images plays a vital role in land cover monitoring, urban management, and civil engineering. Aerial imagery segmentation, as a specific semantic segmentation task that assigns a semantic category to each image pixel, has been widely leveraged in HSR remote sensing images understanding since it can provide semantic and location information for objects of interest.

\begin{figure}
    \centering
    \includegraphics[width=0.8\linewidth]{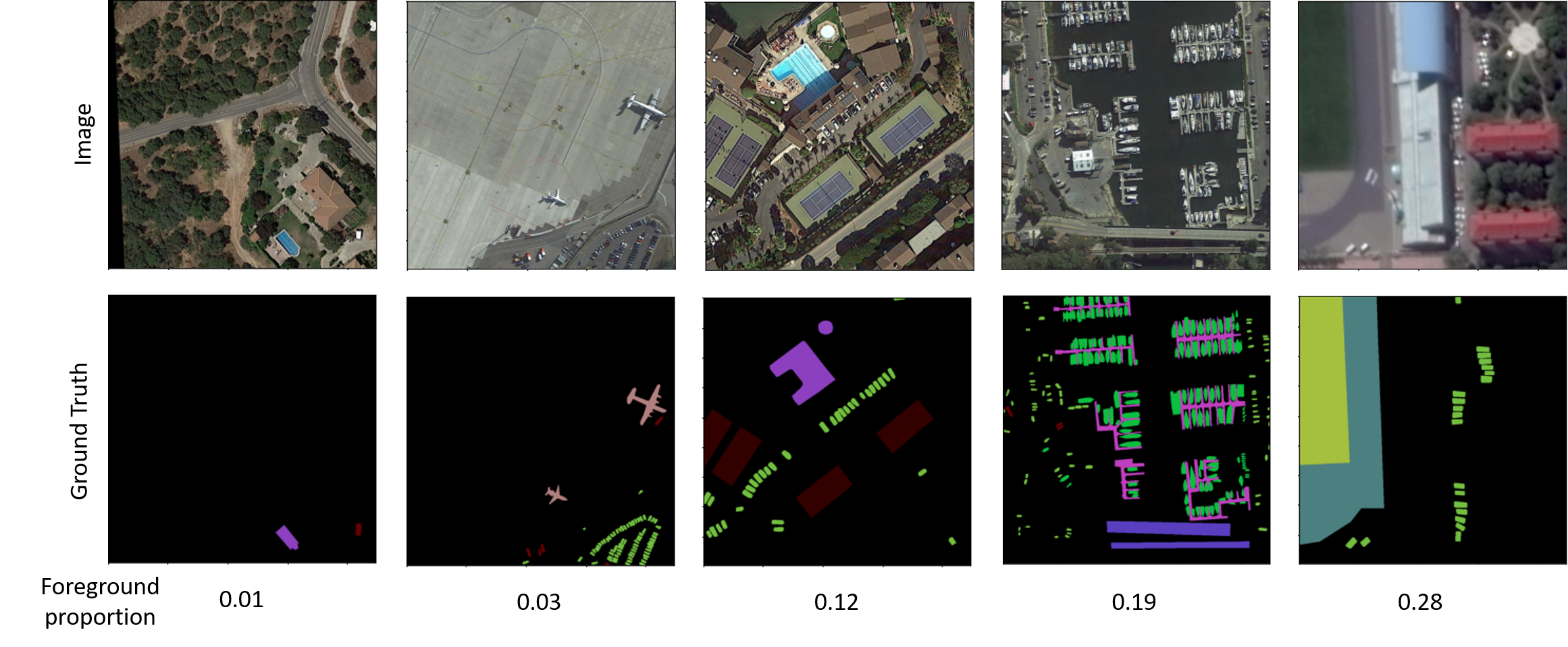}
    \caption{Illustration of aerial imagery segmentation.}
    \label{fig:Show case of aerial image}
\end{figure}

Nonetheless, in addition to some common issues in most semantic segmentation datasets \citep{coco-2018,cityscapes-2016,adek20k-2019}, including large scale variation \citep{PanopticFPN-2019cvpr,FCN2015-cvpr,U-net2015}, complex scene \citep{deeplabv3-2017-cvpr,pspnet(PPM)2017-cvpr}, and indistinguishable object boundaries \citep{cascadepspRefineNet-2020cvpr,Pointrend-2020cvpr,joint-boundary-semantic-Iterative-2020cvpr}, aerial imagery segmentation has its own challenges, the most critical one among which lies in foreground-background imbalance \citep{aerial-problem-2019-1,aerial-problem-2018-1,Pointflow2021-cvpr,aerial-problem-2019-2,iSAID,dota,FarSeg2020-cvpr}. Taking images in Fig.~\ref{fig:Show case of aerial image} as examples, the foreground proportion can be extremely small, e.g. less than 1\% for the leftmost image. Such acute imbalance could drastically increase the difficulty of object recognition, as even human eyes can hardly recognize them from the image. Moreover, the larger intra-class variance of the background objects may significantly increase the risk of false positive results~\citep{Pointflow2021-cvpr,FarSeg2020-cvpr}.

Existing general semantic segmentation methods mainly pay attention to designing sophisticated neural network architectures that can obtain informative multi-scale feature representations \citep{deeplabv3p-2018-eccv,resnet2016-cvpr,PanopticFPN-2019cvpr,hrnet2019,inceptionv4-2017} and highlight the object boundaries  \citep{Pointrend-2020cvpr,Pointflow2021-cvpr,joint-boundary-semantic-Iterative-2020cvpr}. To further address the foreground-background imbalance challenge of aerial imagery segmentation, some recent efforts have created more delicate modules in the neural networks to obtain superior results. For instance, Foreground-Aware Relation Network (FarSeg) \citep{FarSeg2020-cvpr} introduces a foreground-scene relation module to enhance the discrimination of foreground features as well as a foreground-aware optimization to alleviate foreground-background imbalance problem. More recently, PointFlow \citep{Pointflow2021-cvpr} designs a dual point matcher to select points from the salient area and object boundaries.

While these previous studies have demonstrated the effectiveness of sophisticated neural network architecture design in both general and aerial imagery segmentation, most of them ignore the other side of the coin, i.e. how to efficiently utilize the multi-scale representations generated by complex neural networks. Intuitively, to identify the semantic meaning of a large object, it is more important for the model to leverage the representations obtained by wider receptive fields since they can represent the semantics of the whole large object rather than confusing the discrimination with interior details. On the other hand, for small objects, it is more efficient for the model to exploit the representations obtained by more concentrated receptive fields because they can focus on the discrimination without being distracted by noisy context. Unfortunately, most of the existing general or aerial segmentation methods merely employ multi-scale representations in ad-hoc ways, either simply concatenating them or arbitrarily using the final layer. This inevitably limits the potential of aerial imagery segmentation.

\begin{figure}
    \centering
    \includegraphics[width=0.6\linewidth]{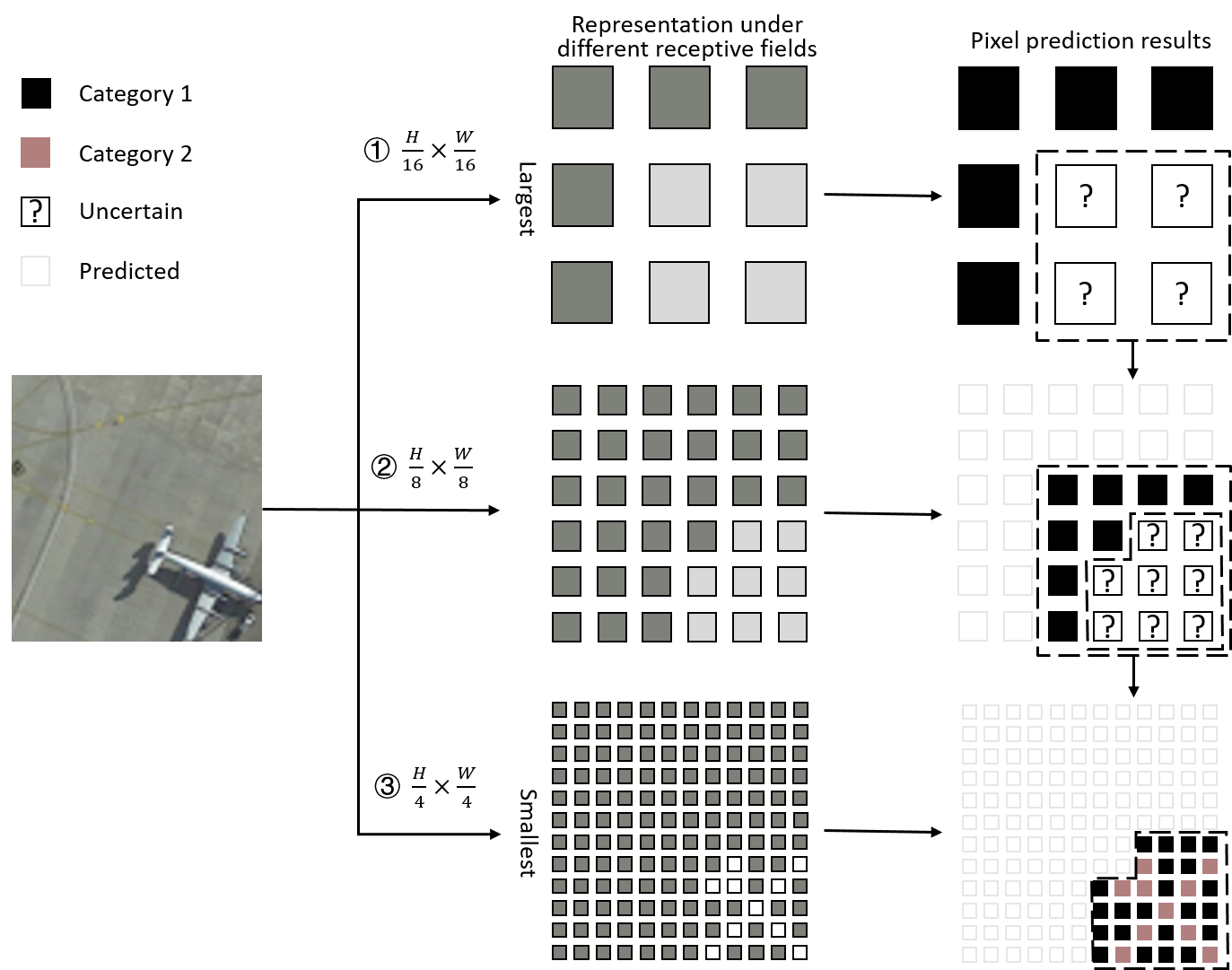}
      \caption{Diagrammatic representation of Adaptive Focus Framework.}
    \label{fig:Demo Framework of Adaptive Focus}
\end{figure}

Inspired by the adaptively-focusing process of the human eye \citep{humaneye-autofocus=2006,autofocus-wikipedia}, we propose a novel Adaptive Focus Framework (AF$_2$) in this paper, which adopts a hierarchical segmentation procedure for aerial imagery segmentation. The general idea of this framework is shown in Fig.~\ref{fig:Demo Framework of Adaptive Focus}. Particularly, through any widely employed model structure (e.g. encoder-decoder structure), AF$_2$ can obtain hierarchical representation maps based on different levels of receptive fields. The semantic segmentation procedure starts from the representation map created by the largest receptive field. After obtaining the segmentation result on this level, AF$_2$ will filter out the pixels of low confidence on segmentation and then carry out another round of segmentation procedure on a finer-grained feature map. The whole process will be conducted repeatedly until all the pixels have been predicted, or until there is no finer-grained feature map. 
In AF$_2$, an Adaptive Confidence Mechanism (ACM) is proposed to deal with pixel filtration. The confidence of a pixel is defined as the highest value among the probabilities that the pixel belongs to each category, while an adaptive updated threshold is set to filter out low confidence pixels.

% Inspired by the focusing process of the human eye and the camera, we propose a novel Adaptive Focus Framework (AF$_2$) in this paper, which adopts a hierarchical segmentation procedure for each aerial imagery. The general idea of this framework is shown in Fig.~\ref{fig:Demo Framework of Adaptive Focus}. As we can see, through the encoder-decoder structure, AF$_2$ can obtain a sequence of feature maps with different levels of receptive fields. The semantic segmentation procedure starts from the feature map with the largest receptive field. For the segmentation results obtained in this level, AF$_2$ filters out the pixels with high uncertainty, and then carries out segmentation procedure on a finer-grained feature map for 
% them. This process is repeated until all the pixels have been predicted, or there is no finer-grained feature map. 
% % As we can see, the filtration in each step is an important part. 
% In AF$_2$, an Adaptive Confidence Mechanism (ACM) is proposed to deal with pixel filtration. The confidence of a pixel is defined as the highest value among the probabilities that it belongs to each category. In which, an adaptive updated threshold is set to filter out low confidence pixels. 

In summary, the main contributions of this paper include: 
\begin{itemize}
\item We propose Adaptive Focus Framework (AF$_2$), which adopts a hierarchical segmentation procedure and focuses on adaptively utilizing multi-scale representations generated by widely adopted neural network architectures.
\item In the pixel filtering process for each scale's representation, the Adaptive Confidence Mechanism (ACM) is adopted to dynamically decide which pixels need to use finer-grained features. This mechanism ensures the performance and robustness of the framework.
\item  Extensive experiments and analysis demonstrate the advantage of AF$_2$. It has significantly improved the accuracy on three typical aerial benchmarks. At the same time, its convergence speed and inference speed are also the fastest.
\end{itemize}
It is worth noting that AF$_2$ is a general framework, focusing particularly on efficient representation utilization, which can be leveraged to benefit any other semantic segmentation task.

% In summary, the main contributions of this paper include following parts: 
% \begin{itemize}
% \item We propose a novel hierarchical segmentation approach without model structure re-design for aerial imagery segmentation, called Adaptive Focus Framework (AF$_2$), which segments the object with the appropriate receptive field.
% \item In the pixel filtering process for each level, the Adaptive Confidence Mechanism (ACM) is adopted to dynamically decide which pixels need further processing. This mechanism ensures the performance and robustness of the framework.
% \item  Extensive experiments and analysis indicate the advantage of AF$_2$. It has significantly improved the accuracy on three typical aerial benchmarks. At the same time, its convergence speed and inference speed are also the fastest.
% \end{itemize}
% AF$_2$ is a methodological innovation based on efficient representation utilization. It has the potential to benefit the general semantic segmentation.

\section{Related Works}

\subsection{General Semantic Segmentation}

Fully-convolutional networks (FCNs~\cite{FCN2015-cvpr}) are the earliest method of using deep learning to model semantic segmentation problem. The backbone models~\citep{resnet2016-cvpr,vgg2014,hrnet2019,inceptionv4-2017,inception-rethinking,resnext2017} are used to generate a lower resolution output than the input image and use bilinear up-sampling to recover the original image resolution. 
Employing the dilated convolution~\citep{deeplabv3-2017-cvpr,deeplabv3p-2018-eccv,dilatedConv-2015} to replace the down-sampling operation will have a better performance at the expense of more
memory and computation cost. 
The spatial context information can overcome the limited receptive field of convolution layer to a certain extent such as Atrous Spatial Pyramid Pooling (ASPP~\cite{deeplabv3-2017-cvpr,deeplabv3p-2018-eccv}), Pyramid Pooling Module (PPM~\cite{pspnet(PPM)2017-cvpr}) , Densely connected Atrous Spatial Pyramid Pooling (DenseASPP~\cite{denseaspp-2018cvpr}) , Relation-Augmented fully convolutional network (RA-Net~\cite{RANet-cvpr2019}), etc. 
PointRend~\citep{Pointrend-2020cvpr} performs point-based segmentation predictions at adaptively selected locations based on an iterative subdivision algorithm.
Some other works \citep{li2020improving,yuan2020segfix,zhang2020dynamic} propose architectures specific for segmentation boundary that is difficult to predict.

% \citep{li2020spatial,yuan2019object,zhang2020dynamic}
The encoder-decoder architectures  \citep{deeplabv3p-2018-eccv,PanopticFPN-2019cvpr,FPN2017-cvpr,U-net2015,gatescnn-2019iccv,li2019gff} progressively upsample the high-level features and combine them with the features from lower levels, ultimately generating high-resolution features. For instance, Deeplab v3+ \citep{deeplabv3p-2018-eccv} combines dilated convolutions with an encoder-decoder structure to produce the output on a grid 4x sparser than the input. SemanticFPN \citep{PanopticFPN-2019cvpr} merges the information from all levels of the Feature Pyramid Network (FPN) pyramid into a single output and produces a dense prediction. 

% In addition to traditional architectures such as FCN and encoder-decoder, there has been efforts in neural architecture search and dynamic networks dedicated for semantic segmentation. \cite{chen2018searching} constructs a recursive search space to search for multi-scale modules specific for dense image prediction. \cite{liu2019auto} conducts joint semantic segmentation architecture search at both the network and cell level. \cite{li2020learning} proposed Dynamic Routing that studies the scale distribution of each image and thus generates a data-dependent forward path.

% These general semantic segmentation methods ignore the special issues including imbalanced foreground-background pixels for modeling the context, increased small foreground objects, huge distribution difference among aerial imageries. 

\subsection{Semantic Segmentation for Aerial Imagery}
\label{sec:Semantic Segmentation for Aerial Imagery}

In recent years, employing deep learning to accelerate the understanding of aerial image has received widespread attention \citep{aerial-image-2017-1,aerial-image-2017-2,aerial-image-2018-1-ISPRS,aerial-image-2018-2-ISPRS,aerial-image-2017-3}, and benefits a lot of applications such as agriculture vision \citep{agriculture-2016,agriculture-2018-1,agriculture-2018}, road extraction \citep{roadextraction-2018,roadextraction-2018-2}, land cover mapping \citep{landcover-2016,landcover-2018-iclr-microsoft,landcover-2019-cvpr-microsoft}, forest monitor \citep{forest-monitor-2019,forest-monitor-2015}, etc. They often design delicate structures to ensure that general semantic segmentation migrates well for specific application scenarios. For instance, Relation Augmented network (RA-Net~\cite{RANet-cvpr2019}) proposes a spatial relation module and a channel relation module to explicitly model global relations. Foreground-Aware Relation Network (FarSeg~\cite{FarSeg2020-cvpr}) enhances the discrimination of foreground features and proposes a balanced optimization based on focal loss \citep{focalloss2017-cvpr}. PointFlow \citep{Pointflow2021-cvpr} designs a dual point matcher to select points from the salient area and object boundaries.

\section{Method}
\label{Method}
\begin{figure}
  \centering
  \includegraphics[width=1\linewidth]{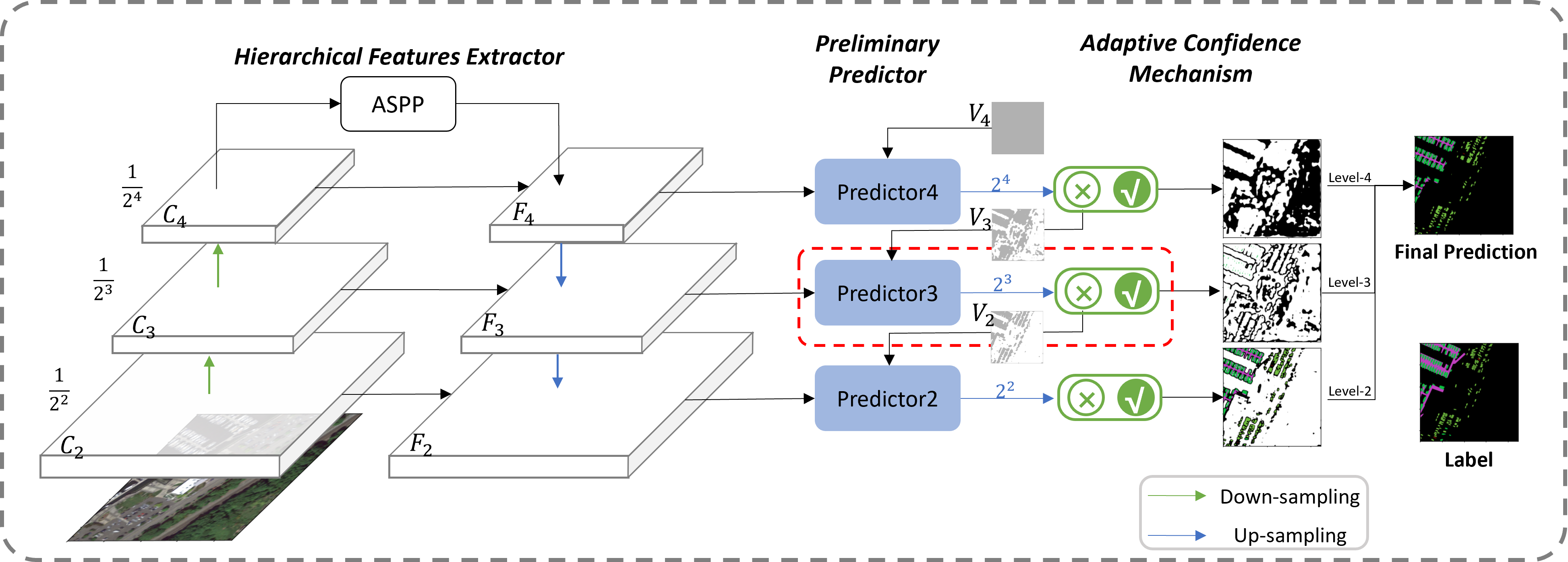}
  \caption{Adaptive Focus Framework. $C_l$ and $F_l$ represent different hierarchy feature maps. $V_l$ means the pixel set fed into level $l$  and we show them in gray. The adaptive confidence mechanism (ACM) is employed to judge whether the prediction for each pixel is sufficiently confident or not in each level, and \checkmark is the sign of confidence. The detail of the red dashed box is shown in Fig. \ref{fig:The process of adaptive confidence mechanism.}.}
  \label{fig:Framework of Adaptive Focus algorithm.}
\end{figure}

In this section, we discuss details of the proposed AF$_2$. As shown in Fig. \ref{fig:Framework of Adaptive Focus algorithm.}, AF$_2$ consists of three main parts: hierarchical features extractor, predictor,  and adaptive confidence mechanism (ACM). The pseudo-code of Adaptive Focus Framework is shown in appendix.

% ~(e.g. $V_4=\{(i,j) \mid i\in [1,H], j\in [1,W]\}$, $H,W$ is the image height and width),

% We propose  Multiple confidence switches are employed sequentially to find desired feature maps to make a precise prediction. In other words, the confidence switch will employ the high-level feature maps to determine whether we need to use the low-level feature maps. In this section, we make a detailed algorithm 
% which can easily and efficiently migrate general semantic segmentation method to aerial imgery scenario.
\begin{itemize}
    \item \textbf{Hierarchical features extractor:} Given an input image, this is used to extract the hierarchical feature maps of the image. The higher the level of feature map, the coarser the granularity of feature map and the larger the receptive field. 
    %Most network based on encoder-decoder can be employed here.  
    %encoder-decoder
    % The hierarchical feature maps are further processed to produce a new set of feature maps that preserves the hierarchical structure but has richer semantic spatial information. Hierarchical semantic predictions are produced based on these feature maps.
    \item \textbf{Preliminary predictor:} For each level of the feature map, there is a preliminary predictor to output the categories of the pixels that still cannot be confidently predicted with higher-level feature maps.   
    \item \textbf{Adaptive confidence mechanism for prediction selection:} The adaptive confidence mechanism (ACM) is employed to judge whether the prediction for each pixel is sufficiently confident or not in each level. The predictions with high enough confidence will be accepted for the corresponding pixels. Otherwise, the pixels will be passed down to the next hierarchy for further prediction with finer-grained features, until there is no pixels for prediction or the lowest level is reached.
    % Each level of the hierarchical predictions has a corresponding confidence threshold to decide whether is sufficiently confident to be contributed into the final result.
    
    % Preliminary prediction
    % Prediction selection with adpative confidence  
    
    % If not, our workflow will go down the hierarchy and check the prediction at the next level that has a finer grain. This procedure is repeated until every position has been confidently predicted or the lowest level feature is reached.
\end{itemize}

%% 讲一下为啥要用 multi-level features,对应不同感受野
\subsection{Hierarchical Features Extractor}
Hierarchical feature extractor is the fundamental part of the AF$_2$. It prepares different levels of feature maps for further pixel-level classification. The higher the level of feature map, the larger the receptive field. Since AF$_2$ is independent of the particular structure of the network, any mainstream feature extraction network can be adopted as long as it is capable of extracting a hierarchy of feature maps. %That is to say, given an image $ I \in \mathbb{R}^{C_{in} \times H \times W} $, the network should outputs a set of feature maps.
% Thus, we select 3 typical model architecture，
% to do 

The typical model structures, such as Fully-Convolutional Network (FCN \cite{FCN2015-cvpr}), Feature Pyramid Networks (FPN \cite{FPN2017-cvpr}) , Semantic FPN \cite{PanopticFPN-2019cvpr}, etc., can be employed here.
In this section, we take FPN \citep{FPN2017-cvpr} with Atrous Spatial Pyramid Pooling module (ASPP) \citep{deeplabv3-2017-cvpr} as an example. Specifically, ResNet \citep{resnet2016-cvpr} is chosen as the backbone network for basic feature extraction. We denote the different level of feature maps generated by ResNet as $\mathcal{C} = \left\{C_l  \big|  C_l\in\mathbb{R}^{d_l\times \frac{H}{2^l} \times \frac{W}{2^l}},\,l\in \left[L_{\min},L_{\max}\right]\right\}$, where $H$ and $W$ represent the image's original height and width. $l$ is the level of feature map, while $L_{\min}$ and $L_{\max}$ are the lowest and highest levels (e.g. if the output stride of ResNet is $2^5$, its $L_{\max}$ is $5$. The lowest feature map's output stride is always $2^2$, and its $L_{min}$ is 2.). Furthermore, let $\mathcal{F} = \left\{F_l  \big|  F_l\in\mathbb{R}^{d_l\times \frac{H}{2^l} \times \frac{W}{2^l}} , l\in \left[L_{\min},L_{\max}\right]\right\} $, which stands for the  the feature map set from the decoder part. $F_l$ is defined as %in the  top-bottom feature fusion is employed like follow:
% The hierarchical feature extractor outputs a set of feature maps with different receptive fields and the whole procedure can be formulated as follows:
\begin{equation}
    F_l = \left\{ \begin{array}{ll}
         f\left(\textsc{ASPP}(C_l),C_l  \mid \theta_l \right),&l=L_{\max} \\
         f\left(\textsc{up}_2(F_{l+1}),C_l \mid \theta_l \right), & \text{otherwise}.
    \end{array}\right.
\end{equation}
where $f(\cdot,\cdot\mid \theta_l)$ represents the top-down feature fusion function. $\textsc{ASPP}$ represents Atrous Spatial Pyramid Pooling Module \citep{deeplabv3-2017-cvpr}. $\textsc{Up}_2$ represents the bilinear up-sampling function with a scale factor $2$. 

\subsection{Preliminary Predictor}
Multiple pixel-level preliminary predictors are assigned for the different feature maps, and the goal of the predictor is to output the category of the pixels that cannot be confidently predicted with higher-level features.
Specifically, the feature map $F_l$ is fed into its corresponding predictor for level $l$ to obtain the prediction results. Since the results belong to $\mathbb{R}^{n\times \frac{H}{2^l} \times \frac{W}{2^l}}$ ($n$ is the category number), the bilinear up-sampling method is employed to generate a prediction with the same size as the original image. The whole process can be formulated as\footnote{Certainly, $P_l =\textsc{Softmax}\left( f\left(\textsc{Up}_{2^l}\left(F_l\right) \mid \theta\right)\right)$ is another option which performs up-sampling first.}:
\begin{equation}
    P_{l} = \textsc{Softmax}\left(\textsc{Up}_{2^l}\left(g\left(F_l \mid \theta_l\right)\right)\right),
    \label{eqn:predictor form}
\end{equation}
where $g(\cdot\mid \theta_l)$ is the prediction function based on multi-layer perceptron (MLP), and $\textsc{Up}_{2^l}$ is the bilinear up-sampling method with a scale factor $2^l$. 

In fact, only a part of the prediction results will be selected from $P_l$.
Specifically, we denote $p_{l,i,j}$ as the prediction results vector in $P_l$ for pixel $(i,j)\in V_l$, where $V_l$ is the pixel set in the original image that still cannot be confidently predicted in higher levels like Eq. \eqref{eqn:pixel set generation}. Only these selected prediction results, i.e. $\left\{p_{l,i,j} \mid (i,j)\in V_l\right\}$, will be processed by ACM to determine whether they are adopted as the final results.

% Based on this fact, we can only use the neighbor pixels of $(i,j)\in V_l$ involved in the bi-linear interpolation to accelerate the process of prediction in implementation.

% \begin{equation}
%     p_{l,i,j}=P_{l,i,j}, ~ (i,j)\in V_l
% \end{equation}

% Particularly, we feed the feature map $F_l$ and pixel set $V_l$, that is the pixels still can not be predicted in higher-levels, into predictor of level $l$. More details about $V_l$ can be found in Section \ref{sec:Adaptive Confidence Switch}. As the $F_l$ belongs to $\mathbb{R}^{d_l\times \frac{H}{2^l} \times \frac{W}{2^l}}$ and the index of pixel set $V_l$ is from original image height/width. Thus, the mathematical formalism is like Eq. \eqref{eqn:predictor form}. In fact, except the highest level predictor, the predictor can only see partial pixels of the image, and multiple predictors can be thought as a delicate boosting process with the assistance of adaptive confidence mechanism.

%todo 丰富一下动机 对下面的first 和second 的intuition动机
\subsection{Adaptive Confidence Mechanism for Prediction Selection}
\label{sec:Adaptive Confidence Switch}
\begin{figure}
  \centering
  \includegraphics[width=1\linewidth]{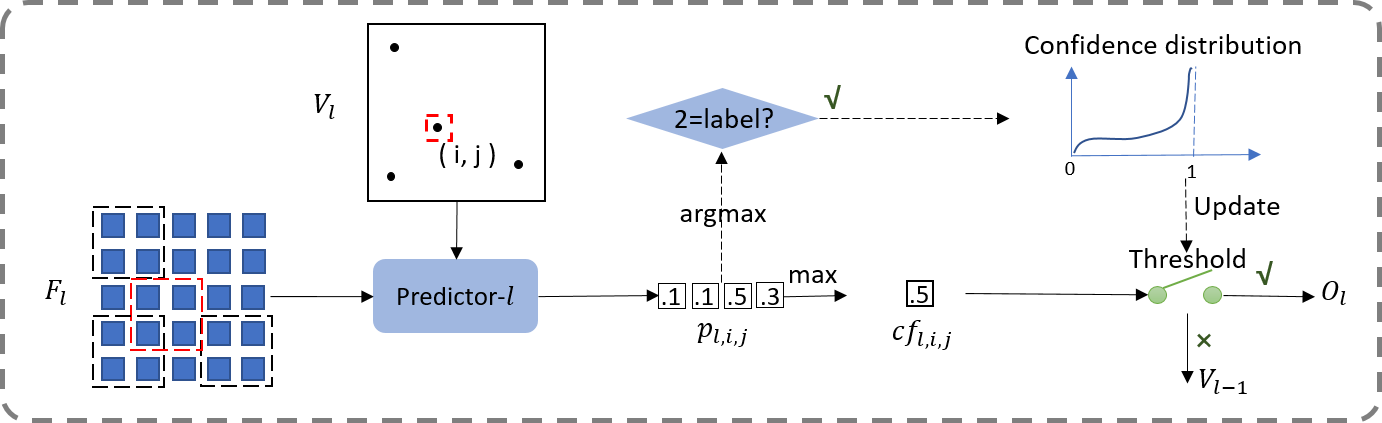}
  \caption{The process of Adaptive Confidence Mechanism. Firstly, the corresponding feature map of pixels in $V_l$ are fed into predictor $l$ to get predicted probabilities. Then, the threshold is employed to judge whether the pixel should go down the hierarchy to get lower-grained features or output the prediction. 
  Meanwhile, the threshold is periodically updated according to the confidence distribution of correctly predicted pixels.
  Note that the threshold will be fixed without update during the inference.}
  \label{fig:The process of adaptive confidence mechanism.}
\end{figure}

The purpose of adaptive confidence mechanism (ACM) is to determine which pixels can be identified and which pixels need to be fed into the lower level for prediction.
To achieve the above, the metric for each pixel is defined and is named as the pixel's confidence. Then, the filtration function is adopted based on this metric.
The process of ACM is depicted in Fig.~\ref{fig:The process of adaptive confidence mechanism.}.

%Intuitively, pixels located in the area nearby multiple categories are hard to predict and may need more detailed information. In contrast, pixels surrounded by a same category as themselves are easy to predict.

%As shown in Fig. \ref{fig:The process of adaptive confidence mechanism.}, the core of adaptive confidence mechanism (ACM) is to filter the pixel easy to classify with current level feature maps, and others (hard pixels) need to use lower-grained feature maps.
%Sequentially execute upper process. 
 %In this section, we make a detailed introduction about the mechanism of adaptive confidence.
% Based on above-mentioned, we develop the definition of confidence for every pixel, the mathematical formalism of ACS. And then, we introduce the Adaptive Confidence Switch process in detail.

% threshold
%% todo,加一点confidence是怎么生成的

% Supposed that, in level $l$, there are $n_l$ pixels with corresponding feature maps are feed into $l$th predictor, and we name the set of these $n_l$ pixels as $V_l$.
Firstly, the pixel's confidence is defined to evaluate whether the corresponding prediction result is reliable enough (i.e. the larger the pixel's confidence, the more reliable the prediction). For the pixel $(i,j)\in V_l$, the  prediction result $p_{l,i,j}$ is an $n$-dimensional vector to indicate the probabilities belonging to the $n$ categories. We define the highest value among these probabilities as its confidence\footnote{The probability distribution entropy, the difference value between the first and second largest probability value, etc. are other optional choices for pixel's confidence.}, that is $ 
    cf_{l,i,j} = \max\left(p_{l,i,j}\right),(i,j)\in V_l.$
%where, $cf_{l,i,j}$ is the confidence value for pixel $(i,j)$ in level $l$. $p_{l,i,j}\in \mathbb{R}^n$ and $n$ is the category number.

Secondly, we filter out some of the pixels to be fed into the lower level according to their confidence. A straightforward way is to select the pixels with the top $k$-lowest confidence in each image. However, since the variance among different images, especially for aerial images, is very large, the top-$k$ approach cannot efficiently handle such complicated situation. For instance, pixels in complicated images usually have low confidence, while pixels in simple images have relatively higher confidence.
Therefore, we choose to use a uniform threshold for all images. Since the accuracy of prediction is improved during training, the threshold should also be learnable and adaptively updated with model training.
% and adapted to the model update (todo). 
%The pixel with a correct prediction can make the confidence bigger and bigger, which is more stable. Meanwhile, the pixel with a 
%wrong prediction need to adjust the $\arg\max$ index and the confidence maybe not stable.   
More concretely, the statistical information of the confidence for correctly predicted pixels in each level is employed to help update the threshold adaptively. 
That is
\begin{equation}
    \tau_l^{t+1} = \gamma \cdot \tau_l^{t} + 
            (1-\gamma) \cdot \textsc{Quantile}_{r}
                \left\{cf_{l,i,j}\mid (i,j)\in V_l\land \arg\max(p_{l,i,j}) = label_{i,j}\right\},
    \label{eqn:threshold soft update function}
\end{equation}
where $\tau_l^t$ is the threshold in step $t$ for level $l$, $\gamma$ is soft update factor, 
%the parameter that controls the intensity of the soft update (default set 0.9). 
and $\textsc{Quantile}_{r}$ represents the $r$-quantile method. % ($0\% \leq r \leq 100\%$), and $r$ is a hyper-parameter. 
It should be noted that the threshold is fixed without update in inference.

Thus, compared with threshold $\tau _l$, the pixels with a lower confidence will be fed into the level $l-1$ like Eq. \eqref{eqn:pixel set generation} except for the lowest level. Meanwhile, the categories of the rest pixels will be determined based on the prediction results of this layer like Eq. \eqref{eqn:final prediction in different level}. In the lowest level $L_{min}$, we set $\tau_{L_{min}}=0$ so that all the pixels $V_{L_{min}}$ will be determined. %since there is no lower level
\begin{eqnarray}
    V_{l-1} =  \{(i,j) \mid (i,j)\in V_l\land cf_{l,i,j}<\tau_l^t\}, l\neq L_{min},
    \label{eqn:pixel set generation}\\
    O_l = 
    \{(i,j,\arg\max(p_{l,i,j})) \mid (i,j)\in V_l\land cf_{l,i,j}\geq \tau_l^t\},
    \label{eqn:final prediction in different level}
\end{eqnarray}
where $O_l$ is the confident prediction in each level.

%% todo 

% In ACM, we can compare pixel's confidence $cf_{l,i,j}$ with the corresponding threshold to get the current level output. Note that, in the lowest level, we do not need to compare the confidence with the threshold as we can not go down the feature maps. Like we mentioned above, if the confidence is lower than threshold, we will go down the hierarchy to get lower-grained features. 
% \begin{equation}
%     O_{l,i,j} = \left\{\begin{array}{ll}
%         p_{l,i,j}, & (i,j)\in V_l,  ~l\in [L_{\min}+1,L_{\max}],~cf_{l,i,j}\geq ts_l^t  \\
%         None, & (i,j)\in V_l,       ~l\in [L_{\min}+1,L_{\max}],~cf_{l,i,j}< ts_l^t \\
%         p_{l,i,j}, & (i,j)\in V_l,  ~l=L_{\min} \\
%         None, & (i,j)\notin V_l ,
%     \end{array}\right.
%     \label{eqn:the output in level l}
% \end{equation}

Since there are a number of pixels whose categories could be determined in each level, i.e., $O_l$ for $l_{\min} \leq l \leq l_{\max}$, the final prediction results is a union of all these pixels $O=\bigcup_{l=L_{\min}}^{L_{\max}}O_l$.

\subsection{Optimization}

The optimization objective function is based on cross-entropy loss. Due to the hierarchical prediction procedure adopted in AF$_2$, the overall loss is accumulated at each level of the hierarchical predictions. Since pixels in $V_l$ are fed into predictor in level $l$, the total loss function can be formulated as follow: 
\begin{equation}
    J = \sum _{l\in [L_{\min},L_{\max}]} \left( 
            \frac{1}{||{V}_l||} \sum _{(i,j)\in V_l}\textsc{Cross-Entropy}\left(p_{l,i,j},label_{i,j}\right)
    \right).
\end{equation}

\section{Experiments}
To evaluate the proposed method, we carry out comprehensive experiments on three aerial imagery datasets: iSAID, Vaihingen and Potsdam. We first introduce these datasets. Then, we show the experimental results on those four datasets. After that, we conduct further analysis to examine the importance of each components of AF$_2$. The implementation details are shown in appendix.

\subsection{Datasets}
\label{Datasets and Implementation Details}
% \textbf{Datasets.} Here we choose three commonly used benchmarks to evaluate the effectiveness of our proposed method. 

\textbf{iSAID.} iSAID~\citep{iSAID,dota} is the largest dataset for instance segmentation in the HSR remote sensing imagery. It contains 2,806 high-resolution aerial images with 655,451 instance annotations from 15 categories. iSAID is distinguished from other semantic segmentation datasets for its significant imbalance between the annotated instances and background as well as its scale variation even for the instances from the same category. In our experiments, we follow its default dataset split where 1,411 images are used for training, 458 for validation, and 937 for testing. The original images can be as large as 4000$\times$13000 pixels. Following previous work~\citep{FarSeg2020-cvpr,Pointflow2021-cvpr}, these images are cropped into patches with a fixed size of 896$\times$896 with a sliding window striding 512 pixels, and these models are trained with 16 epochs on cropped images for all experiments. 
We employ the mean intersection over union (mIoU) as evaluation metric.

% We only use iSAID's semantic masks and did not make any distinguish between the instances. 

\textbf{Vaihingen and Potsdam.} Vaihingen includes 33 aerial images with 2494$\times$2064 pixels. Potsdam includes 38 aerial images with 6000$\times$6000 pixels. 6 categories are defined for both of them. Following the previous work \citep{Pointflow2021-cvpr}, images are cropped into patches with fixed sizes of 768$\times$768 and 896$\times$896, respectively.
These models are trained with 200 epochs for all experiments. We use mIoU and mean-$F_1$ metrics to evaluate the proposed method.

\subsection{Result Comparison}
In this section, we conduct detailed experiments on iSAID, Vaihingen and Potsdam datasets to compare the final prediction performance and the inference efficiency of AF$_2$ with several mainstream methods. 
The base model for hierarchical feature extractor adopted is a combination of FPN \citep{FPN2017-cvpr} and ASPP \citep{deeplabv3p-2018-eccv}, abbreviated as AFPN for convenience. Our whole implementation is denoted as AF$_2$-AFPN.

\textbf{Result Comparison on iSAID.} As shown in Table.~\ref{tab:Experiments on 3 aerial imagery datasets}, AF$_2$-AFPN achieves the state-of-the-art result among all previous works with $r=0.32$ in ACM. The results also show a significant improvement among some categories, which demonstrates again that AF$_2$ can identify the tiny instance with high accuracy. 
Meanwhile, compared with previous methods, It shows that AF$_2$-AFPN has exceeded most mainstream models in terms of running efficiency.
More details can be found in the appendix. 

It is worth noting that the architecture re-design methods, such as FarSeg~\citep{FarSeg2020-cvpr} and the best model before PointFlow~\citep{Pointflow2021-cvpr}, can be easily integrated with our model to further improve the performance. However, tweaking the architecture is not the focus of this work. Meanwhile, the source code of some models, e.g. PointFlow, is not available yet. As one piece of future work, we will verify the effectiveness of the combination of AF$_2$ and other more advanced architectures.

\begin{table}[]
\renewcommand\arraystretch{1.1}         %表格内部 2 倍行距离
\renewcommand\tabcolsep{4.0pt}
    \caption{Experiments on 3 aerial imagery datasets. For iSAID $val$ set, we show the mIoU score and the category with a significant improvement, such as: baseball court (BC), large vehicle (LV), helicopter (HC), swimming pool (SP) and roundabout (RA). All the experiments use ResNet-50 with weights pretrained on ImageNet as backbone for fair comparison except HRNet \citep{hrnet2019}. }
    \label{tab:Experiments on 3 aerial imagery datasets}
    \centering
    \scalebox{0.75}{
    \begin{tabular}{c|cccccc|cc|cc}
\toprule
\multirow{2}{*}{Method}&   \multicolumn{6}{c|}{iSAID (\%)}  & \multicolumn{2}{c|}{Vaihingen (\%)}  & \multicolumn{2}{c}{Potsdam (\%)} \\
 &  mIoU & BC & LV & HC & SP & RA & mIoU & m-$F_1$ & mIoU & m-$F_1$\\ \hline
FCN \citep{FCN2015-cvpr}  & -  & - & - & - & - & - & 64.2  & 75.9  & 73.1  & 83.1 \\
PSPNet \citep{pspnet(PPM)2017-cvpr}  & 60.3    & 61.1  & 58.0  & 10.9  & 46.8  & 68.6  & 65.1  & 76.8  & 73.9  & 83.9 \\
Ocnet \citep{OCNet2018yuan}  & - & -  & - & - & - & - & 65.7  & 77.4  & 74.2  & 84.1 \\
DenseASPP \citep{denseaspp-2018cvpr}  & 57.3    & 54.8  & 55.6  & 33.4  & 37.5  & 53.4  & 64.7  & 76.4  & 73.9  & 83.9 \\
Deeplabv3+ \citep{deeplabv3p-2018-eccv}  & 61.5    & 56.6  & 60.3  & 34.5  & 41.4  & 65.1  & 64.3  & 76.0  & 74.1  & 83.9 \\
SemanticFPN \citep{PanopticFPN-2019cvpr}  & 62.1    & 54.1  & 61.0  & 37.4  & 42.8  & 70.2  & 66.3  & 77.6  & 74.3  & 84.0 \\
RefineNet \citep{cascadepspRefineNet-2020cvpr}  & 60.2    & 61.1  & 58.2  & 23.0  & 43.4  & 65.6  & - & - & - & -\\
UPerNet \citep{UPerNet-2018-jointLearn-multitask}  & 63.8    & 55.3  & 61.3  & 30.3  & 45.7  & 68.7  & 66.9  & 78.7  & 74.3  & 84.0 \\
HRNet \citep{hrnet2019}  & 61.5  & 59.4  & 62.1  & 14.9  & 44.2  & 52.9  & 66.9  & 78.2  & 73.4  & 83.4 \\
GSCNN \citep{gatescnn-2019iccv} & 63.4  & 56.1  & 63.8  & 33.8  & 48.8  & 58.5  & 67.7  & 79.5  & 73.4  & 84.1 \\
SFNet \citep{SFNet-eccv-2020}  & 64.3  & 58.8  & 62.9  & 30.4  & 47.8  & 69.8  & 67.6  & 78.6  & 74.3  & 84.0 \\
RANet \citep{RANet-cvpr2019}  & 62.1  & 53.2  & 60.1  & 38.1  & 41.8  & 70.5  & 66.1  & 78.2  & 73.8  & 83.9 \\
PointRend \citep{Pointrend-2020cvpr}  & 62.8  & 55.4  & 62.3  & 29.8  & 45.0  & 66.0  & 65.9  & 78.1  & 72.0  & 82.7 \\
FarSeg \citep{FarSeg2020-cvpr}  & 63.7  & 62.1  & 60.6  & 35.8  & 51.2  & 71.4  & 65.7 & 78.0 & 73.4 &83.3 \\
PointFlow \citep{Pointflow2021-cvpr}  & 66.9  & 62.2  & 64.6  & 37.9  & 50.1  & 71.7  & 70.4 & 81.9 & \textbf{75.4} & \textbf{84.8}\\
AF$_2$-AFPN  & \textbf{67.8}  & \textbf{66.2}  & \textbf{67.3}  & \textbf{38.9}  & \textbf{53.1}  & \textbf{77.0}  & \textbf{70.5}  & \textbf{82.1}  & 74.9  & 84.4 \\

\bottomrule
    \end{tabular}}

\end{table}

\textbf{Result Comparison on Vaihingen and Potsdam.}
% \subsection{Results on Vaihingen and Potsdam}
To further verify the performance of our framework, we also conduct experiments on two well-known aerial imagery datasets, i.e. Vaihingen and Potsdam. Compared with iSAID dataset, the categories are relatively balanced in terms of instance shape and category ratio in these two datasets. 
% This is particularly evident in Potsdam dataset, and leads to unobvious gaps between different methods. 
The quantitative results listed in Table~\ref{tab:Experiments on 3 aerial imagery datasets} show that AF$_2$ outperforms all methods except PointFlow. More details can be found in the appendix.

\subsection{Sensitivity Analysis}
\begin{figure}[]
  \centering
  \includegraphics[width=0.9\linewidth]{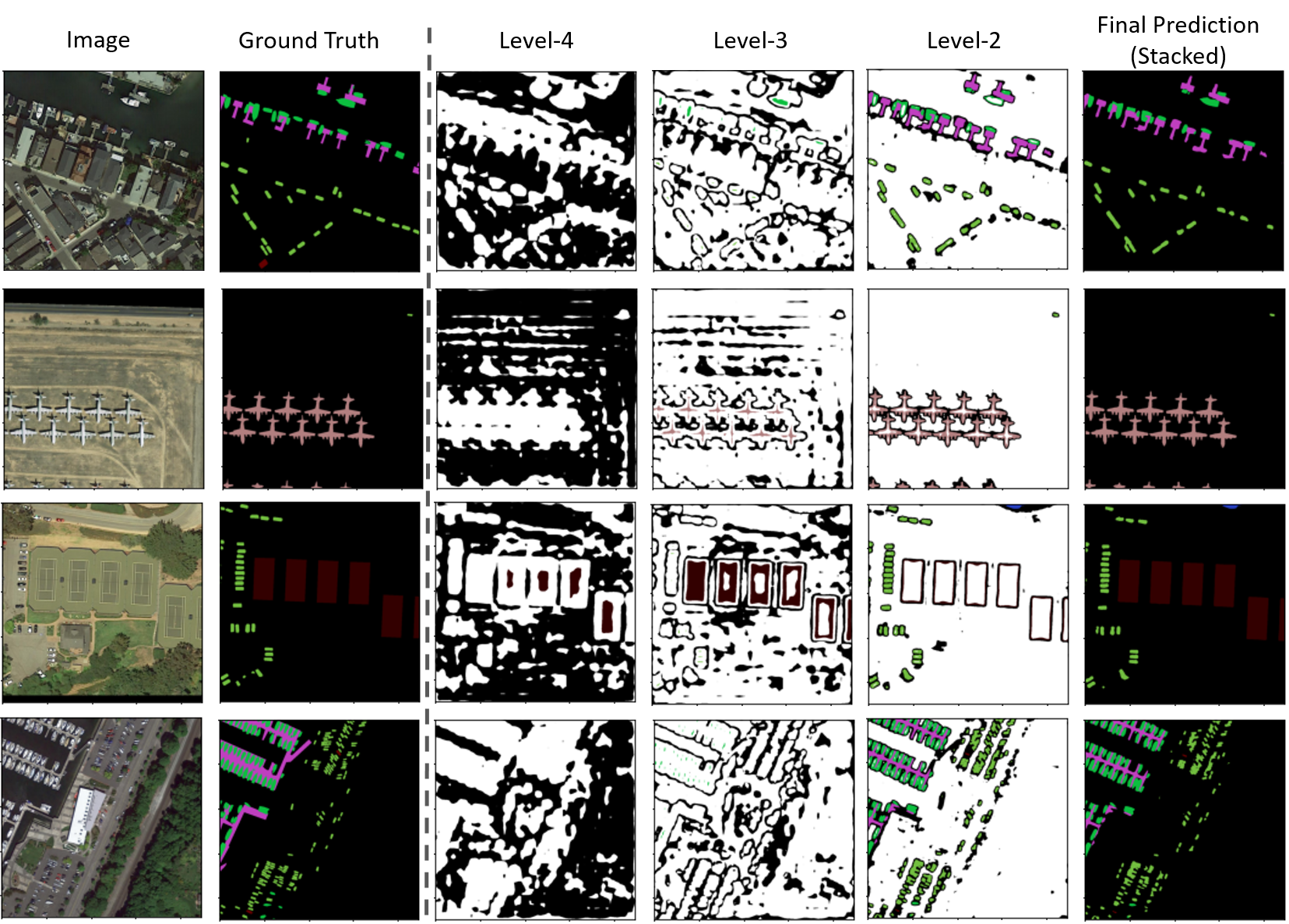}
  \caption{Image prediction process show case on iSAID validation set with $r=0.32$. Different color means different category. Black means the background. Specifically, white means that the pixel has not been predicted or has been predicted in higher level.}
  \label{fig:Image Prediction Process Show Case}
\end{figure}

In this section, we conduct thorough analysis over iSAID to study the importance of each modules of AF$_2$.
%\paragraph{Show Case of Image Prediction Process.} 
To better demonstrate the analysis results, we first present the process of image prediction over some examples from the validation set of iSAID in Figure~\ref{fig:Image Prediction Process Show Case}.
%We show the detailed prediction process of some images in iSAID dataset to illustrate the working mechanism of AF$_2$. 
In this figure, columns 3 to 5 indicate the accepted prediction results at level 4 to 2, respectively, while the last column is the stacked final results. As we can see, at higher levels, it is more often the inner areas of the background or foreground that are predicted successfully. Lower level finer-grained features are primarily used to improve the prediction results of the edges. 
%the big instances are mostly predicted in the high level with the coarse features, while the tiny instance and the boundary benefit from the fine features. 
This hierarchical prediction method is independent to the proportion of foreground and background and thus can effectively solve the problems existed in the previous method.
%Meanwhile, the prediction ratio at different levels is totally different among images, which indicates that our method can handle the category distribution well.

\textbf{Study on the Effect of the ACM.} ACM, as a core part of AF$_2$, is designed to filter out the low confidence pixels. In ACM, the quantile ratio is a key factor to control the filtration threshold. To study its influence, we set different values of $r$ in ACM.
As shown in Table~\ref{tab:ablation study for quantile method}, the models integrated with AF$_2$ achieve significant improvement over the basic AFPN model. 
In addition, the experimental results show that the setting of $r$'s value has a great influence on the results. 
Particularly, either too large or too small $r$ will skew the model toward finer- or coarser- grained representations. Experimental results show that 0.3 is a compromise value on the iSAID dataset, which can greatly improves the prediction accuracy over the baseline method. It is worth noting that, in the case of $r=0.9$, where most pixels (96.0\%) are segmented in level-2, the mIoU value has been improved by 2.1\%. This result indicates that the loss function for coarse-grained features is beneficial.

% shows the importance of coarse-grained features.  

%when the $r$ is big enough (e.g. $r=0.9$), although all the prediction happens in level-2 with the finest-grained feature maps, our method still have an improvement compared with baseline as the loss for multi-levels features.
%As the $r$ decreases, more and more background pixels are predicted in the high level. Meanwhile, the tiny instance and the boundary are predicted in the low level, thus, the low-level predictor has a better prediction with a balanced foreground/background, which strongly demonstrates the explainability and effectiveness of our method.
%Our method has a significant improvement when $r=0.3$ compared to the baseline.

\begin{table}[htbp]
         \caption{Study of ACM. The prediction ratio means how many pixels have completed prediction at this level (i.e. {\footnotesize$\frac{\parallel O_l\parallel}{H\times W}$}). The foreground ratio means foreground pixels proportion of the current pixel sets (i.e. {\footnotesize$ \frac{\parallel \{ label_{i,j}~is ~fg. ~\mid~ (i,j)\in V_l\} \parallel }{\parallel \mathbf{V}_l \parallel}$}).\vspace{1mm}}
    \label{tab:ablation study for quantile method}
    \centering
    \scalebox{0.75}{
    \begin{tabular}{ccccccccc}
    \toprule
\multirow{2}{*}{Method} & \multirow{2}{*}{$r$}     & \multicolumn{3}{c}{Prediction Ratio (\%)} & \multicolumn{3}{c}{Foreground Ratio (\%)} & \multirow{2}{*}{mIoU (\%)}\\
      &  & level-4 & level-3 & level-2 & level-4 & level-3 & level-2 \\ \hline
AFPN & -& - & - & 100 & - & - & 3.3 & 63.3 \\
AF$_2$-AFPN & 0.9 & 2.2 & 1.8 & 96.0       & 3.3 & 3.5 & 3.6 & 65.4 \\
AF$_2$-AFPN & 0.7 & 9.7 & 16.3 & 74.0     & 3.3 & 3.8 & 4.3 & 65.8 \\
AF$_2$-AFPN  & 0.5 & 32.7 & 41.2 & 26.0     & 3.3 & 4.7 & 8.8 & 66.4 \\
AF$_2$-AFPN & 0.3 & 73.7 & 20.0 & 6.3      & 3.3 & 8.8 & 27.6 & \textbf{67.4} \\
AF$_2$-AFPN & 0.1 & 91.4 & 7.0 & 1.6      & 3.3 & 22.7 & 41.3 & 64.7 \\
    \bottomrule
    \end{tabular}}
    \vspace{-0.5cm}
\end{table}

\paragraph{Study on the Effect of Employed Levels.} In this experiment, we study the influence on performance when employing feature maps from only a subset of levels in AF$_2$. By default, feature maps from level-2, level-3, and level-4 are employed in the framework to produce the final result. However, as shown in Table\ref{tab:Study on the Effect of Employed Levels}, when using only two levels' feature maps, the segmentation mIoU dropped from 67.4 to 66.8 even in the best case. Note that only employing feature maps from level-2 and level-4 is equivalent to FPN and FCN respectively. Experimental results demonstrate that the multi-scale feature maps utilization is crucial.

\textbf{Hierarchical Features Extractor with Different Neural Architectures.}
In this experiment, we aim to evaluate the extendibility and flexibility of the proposed AF$_2$ in terms of feature extraction. Specifically, we select 3 typical architectures: FCN \citep{FCN2015-cvpr}, SemanticFPN \citep{PanopticFPN-2019cvpr}, FPN+ASPP \citep{deeplabv3-2017-cvpr,FPN2017-cvpr} as the hierarchical features extractor, which are named as FCN, SFPN and AFPN for convenience. For FCN, we combine the feature of each level with all of its higher level feature maps to introduce richer semantic information into lower levels. As shown in Table~\ref{tab:Ablation Study for Hierarchical Features Extractor}, we find that the different architectures have different degrees of score improvement with the assistance of AF$_2$. AF$_2$ narrows the gap among different architectures. 
Even for the the FCN architecture, which only has a bottom-up feature generation, AF$_2$ can achieve about 8.4 improvement on mIoU score. This result further illustrates the extendibility and superiority of AF$_2$.

\begin{table}[htbp]
% \centering
\renewcommand\arraystretch{1.1}         %表格内部 2 倍行距离
\renewcommand\tabcolsep{2.0pt}
\begin{minipage}[b]{0.5\textwidth}
    \caption{Study of Employed Levels. }
    \label{tab:Study on the Effect of Employed Levels}
    \centering
    \scalebox{0.75}{
    \begin{tabular}{ccccc}
    \toprule
    Method & Level-4 & Level-3 & Level-2 & mIoU \\ \hline
AF$_2$-AFPN & \checkmark &  & & 57.9 \\ 
AF$_2$-AFPN &  & \checkmark & & 60.7 \\ 
AF$_2$-AFPN &  & & \checkmark & 63.1 \\ 
AF$_2$-AFPN & \checkmark & \checkmark & & 66.0 \\ 
AF$_2$-AFPN & & \checkmark & \checkmark & 66.7 \\ 
AF$_2$-AFPN & \checkmark & & \checkmark & 66.8 \\ \hline
AF$_2$-AFPN & \checkmark & \checkmark & \checkmark & \textbf{67.4} \\ 
    \bottomrule
    \end{tabular}}
\end{minipage}
% \hspace{20pt}
\begin{minipage}[b]{0.5\textwidth}
    % \caption{Ablation Study for Adaptive Confidence Switch}
    \caption{Study of Hierarchical Features Extractor. \vspace{1mm}}
    \label{tab:Ablation Study for Hierarchical Features Extractor}
    \centering
    \scalebox{0.75}{
    \begin{tabular}{cc}
    \toprule
    Method & mIoU \\ \hline
FCN & 57.9 \\
SFPN  & 62.1\\
AFPN & 63.3 \\ \hline
AF$_2$-FCN &  66.3\\
AF$_2$-SFPN & 66.8\\
AF$_2$-AFPN &  \textbf{67.4} \\ 
    \bottomrule
    \end{tabular}}
\end{minipage}

\end{table}

\textbf{Results on general segmentation benchmark} 
We further verify our approach on general segmentation dataset Cityscapes.
As show in Table \ref{tab:Experiment on Cityscapes validation}, the experiment shows that our method has about 2\% mIoU improvement. The training, validation, testing data is 2975, 500, and 1525 respectively. We only use the fine-data for training. During training, data augmentation contains random horizontal flip, random cropping with the size 768$\times$768. We train a totally 50k iterations with a minibatch 16. We use ResNet-50 with weights pretrained on ImageNet.
% importance of efficient utilization of multi-scale feature maps.

\begin{table}[htbp]
% \centering
\renewcommand\arraystretch{1.1}         %表格内部 2 倍行距离
\renewcommand\tabcolsep{2.0pt}
    \caption{Experiment on Cityscapes validation. }
    \label{tab:Experiment on Cityscapes validation}
    \centering
    \scalebox{0.75}{
    \begin{tabular}{ccccccccccccccccccccc}
    \toprule
Methods  &  mIoU  &  road   &  SW   &  BD   &  wall   &  fence   &  pole  &  TL  &  TS  &  VG   &  terrain   &  sky   &  person   &  rider   &  car   &  truck   &  bus   &  train   &  MT   &  bicycle  \\ \hline
 SFPN  &  0.76   &  0.98   &  0.85   &  0.92   &  0.52   &  0.61   &  0.63   &  0.69   &  0.78   &  0.92   &  0.64   &  0.95   &  0.82   &  0.63   &  0.95   &  0.71   &  0.82   &  0.58   &  0.68   &  0.77  \\ \hline
 AF2-SFPN  &  0.78   &  0.98   &  0.86   &  0.93   &  0.57   &  0.62   &  0.66   &  0.71   &  0.80   &  0.93   &  0.65   &  0.95   &  0.83   &  0.66   &  0.95   &  0.76   &  0.84   &  0.70   &  0.67   &  0.78  \\
     \bottomrule
\multicolumn{20}{l}{ SW, BD, TL, TS, VG, MT represents sidewalk, building, traffic light, traffic sign, vegetation, and motorcycle, respectively.} \\

    \end{tabular}}
    \vspace{-0.6cm}

\end{table}

% performance of different hierarchical feature extractors. We select 3 typical model architecture to evaluate the extendibility and flexibility of our framework.
% As we mentioned in Sec. \ref{sec:Adaptive Confidence Switch}, 
% when the distribution among images is different (e.g. the foreground/background ratio for image 1 and image 2 is 1/99 and 30/70), selecting $\textsc{top}k$ hard pixels may be not friendly. Therefore, in this section, we make a comparison between $\textsc{top}k$ and unified threshold method in Tab. \ref{tab:topk and threshold from adaptive confidence switch}. We find that ******

% \begin{table}[]
% \renewcommand\arraystretch{1.1}         %表格内部 2 倍行距离
% \renewcommand\tabcolsep{2.0pt}
%     \centering
%     \caption{Caption}
%     \label{tab:my_label}
%     \scalebox{0.75}{
%     \begin{tabular}{cccc}
%     \toprule

% Method & ACS & $r$ & mIoU \\ \hline
% baseline & - & -\\
% baseline (w/AF) & $topk$ & 0.2 & -\\
% baseline (w/AF) & $topk$ & 0.3 & 67.75\\
% baseline (w/AF) & $topk$ & 0.4 & -\\
% baseline (w/AF) & $topk$ & 0.5 & 67.10\\
% baseline (w/AF) & threshold & 0.3 & 68.17\\
%     \bottomrule
%     \end{tabular}
%     }
% \end{table}

% \textbf{Future Work: } 
% We find another aerial imagery segmentation method named PoindFlow \citep{Pointflow2021-cvpr}, which proposes a new structure to handle the thorny problem in aerial imagery. Although our framework performs better, this method is not conflict with our work. Thus, in our future work, we will verify the effectiveness of the combination of AF$_2$ and PointFlow.

% 很多大片区域都在high level 完成了分类，进一步佐证了我们方法的有效性和合理性。

\section{Conclusion}
In this paper, we argue that the lack of efficient utilization of multi-scale representations could be a bottleneck for accurate semantic segmentation on HSR aerial imagery, which is characterized by the huge scale variation of objects and the imbalance between foreground and background. We present AF$_2$, i.e. Adaptive Focus Framework, to alleviate this critical but long-standing concern. AF$_2$ is independent of the specific architecture and is capable of adaptively utilizing multi-scale feature representations and producing the final result through the proposed Adaptive Confidence Mechanism. Extensive experiments and analyses have demonstrated its remarkable advantages in boosting segmentation accuracy and have proved its universality on common architectures and datasets. 

% Such advantages are achieved with an outstanding speed-accuracy trade-off while without additional parameters in the network.



\bibliography{iclr2022_conference}
\bibliographystyle{iclr2022_conference}

\clearpage
\appendix
\section{Appendix}
\subsection{Pseudo-code of Adaptive Focus Framework}
Pseudo-code is shown in Alg. \ref{code:Pseudo-code of Adaptive Focus Framework} which summarizes and describes how various stages work in training or testing mode.

\renewcommand{\algorithmicrequire}{\textbf{Input:}}  
\renewcommand{\algorithmicensure}{\textbf{Output:}} 

\begin{algorithm}[h]  
  \caption{Pseudo-code of Adaptive Focus Framework}  
  \begin{algorithmic}[1]
    \Require Image: $X\in R^{c\times H \times W}$
    \Ensure Image segmentation
    \State Initialize: $V_{L_{max}}=\{(i,j) \mid i\in [1,H], j\in [1,W]\}$
    \State Mode: train or test
    \State Feature map from FPN: $\{ F_l  \big|  F_l\in\mathbb{R}^{d_l\times \frac{H}{2^l} \times \frac{W}{2^l}} , l\in \left[L_{\min},L_{\max}\right] \} $
    \State Preliminary predictor result: $P_{l} = Softmax\left(Up_{2^l}\left(g\left(F_l \mid \theta_l\right)\right)\right)$
    \For{$l \in \{L_{max},L_{max-1},...,L_{min}\}$}:
        \For{each $(i,j)\in V_l$}:
        \State select pixel (i,j) 's preliminary prediction from $P_{l}$: $p_{l,i,j}=P_{l,(i,j)}$
        \State calculate the confidence value: $cf_{l,i,j} = \max\left(p_{l,i,j}\right),(i,j)\in V_l$
        \If{$cf_{l,i,j}>\tau_l$ and $l \neq L_{min}$}:
            \State put $(i,j,\arg\max(p_{l,i,j}))$ into $O_l$
        \Else
            \State put $(i,j)$ into $V_{l-1}$
        \EndIf
        
        \If{mode is train}:
            \State Adaptively update the threshold:$\tau_l$
        \EndIf
        
        \EndFor
    \EndFor
    
    \State Final prediction is: $O=\bigcup_{l=L_{\min}}^{L_{\max}}O_l$
    \If{mode is train}:
        \State optimize objective function: $J = \sum_{l} \frac{1}{||V_l||} \sum_{(i,j)\in V_l} CrossEntropy(p_{l,i,j},label_{i,j}) $
    \EndIf
    
    \label{code:Pseudo-code of Adaptive Focus Framework}  
  \end{algorithmic}  
\end{algorithm}

\subsection{Implementation Details} Following~\cite{deeplabv3p-2018-eccv,Pointflow2021-cvpr,RANet-cvpr2019,FarSeg2020-cvpr}, we use ResNet-50 with weights pretrained on ImageNet in all the experiments for fair comparison except for HRNet \citep{hrnet2019}. The output stride of the backbone is adjusted to 16 by setting the convolution stride of the last layer to 1 and employing dilated convolution following DeepLab v3+\citep{deeplabv3p-2018-eccv}. The feature maps from the backbone, i.e. $C_2$, $C_3$, $C_4$, are the outputs of the $conv2\_x$, $conv3\_x$, and $conv5\_x$ of the ResNet-50 respectively. 
All the confidence thresholds are initialized to be 0.5. In the threshold update, we set the soft intensity $\gamma$ to 0.9 and $r$ to 0.3 for $\textsc{Quantile}_{r}$. 
The learning rate decreases from 0.01 to 0.0001 following the poly policy, i.e. $lr_{step}=lr_{init}(1-\frac{step}{max\_step})^{power}$, where $power$ is 0.9.
We employ synchronized SGD over 4 GPUs with each mini-batch containing 16 cropped patches, weight decay of
0.0001 and momentum of 0.9. The synchronized batch normalization is enabled for cross-GPU communication.
Following \cite{Pointflow2021-cvpr,RANet-cvpr2019,FarSeg2020-cvpr}, for iSAID, we adopt data augmentation during the training, which includes horizontal flip, vertical flip, and rotations of 90, 180, and 270 degrees. Note that, for each experimental setting, we run the experiment 5 times and report the average score to remove the influence of randomness.

\subsection{Results on iSAID Dataset}

We show the detailed results on iSAID Dataset in Tab. \ref{tab:iSAID experimence valid set}, and AF$_2$ achieves state-of-the-art for most categories. 
% especially for some tiny objects such as storage tank, baseball court (BC), large vehicle (LV), helicopter (HC), swimming pool (SP) and roundabout

\begin{table}
\renewcommand\arraystretch{1.1}         %表格内部 2 倍行距离
\renewcommand\tabcolsep{2.0pt}
  \centering
  \caption{Experimental results on iSAID val set. The bold values in each column represent the best entries. The category are defined as: ship (Ship), storage tank (ST), baseball diamond (BD), tennis court (TC),baseball court (BC), ground field track (GTF), bridge (Bridge), large vehicle (LV), small vehicle (SV), helicopter (HC), swimming pool (SP), roundabout (RA), soccerball field (SBF), plane (Plane), harbor (Harbor). All the models are trained under the same setting following the FarSeg\citep{FarSeg2020-cvpr} and PointFlow \citep{Pointflow2021-cvpr}.}
  \label{tab:iSAID experimence valid set}
  \scalebox{0.65}{
  \begin{tabular}{c|c|c|cccccccccccccccccc}
    \toprule
\multirow{2}{*}{Method} & \multirow{2}{*}{backbone} & \multirow{2}{*}{mIoU(\%)} & \multicolumn{15}{c}{IoU per category(\%)} \\ 
& & & Ship & ST & BD & TC & BC & GTF & Bridge & LV & SV & HC & SP & RA & SBF & Plane & Harbor\\ \hline
PSPNet \citep{pspnet(PPM)2017-cvpr} & ResNet50 & 60.3 & 65.2 & 52.1 & 75.7 & 85.6 & 61.1 & 60.2 & 32.5 & 58.0 & 43.0 & 10.9 & 46.8 & 68.6 & 71.9 & 79.5 & 54.3\\ 
DenseASPP \citep{denseaspp-2018cvpr} & RenNet50 & 57.3 & 55.7 & 63.5 & 67.2 & 81.7 & 54.8 & 52.6 & 34.7 & 55.6 & 36.3 & 33.4 & 37.5 & 53.4 & 73.3 & 74.7 & 46.7\\ 
Deeplabv3+ \citep{deeplabv3p-2018-eccv} & ResNet50 & 61.5 & 63.2 &  67.8 &  69.9 & 85.3 &  56.6 &  52.9 &  34.2 &  60.3 & 43.2 &  34.5 &  41.4 &  65.1 &  73.8 &  81.0 &  52.3\\ 
SemanticFPN \citep{PanopticFPN-2019cvpr} & ResNet50 & 62.1 & 68.9 & 62.0 & 72.1 & 85.4 & 54.1 & 48.9 & 44.9 & 61.0 & 48.6 & 37.4 & 42.8 & 70.2 & 61.6 & 81.7 & 54.9\\ 
RefineNet \citep{cascadepspRefineNet-2020cvpr} & ResNet50 & 60.2 & 63.8 & 58.6 & 72.3 & 85.3 & 61.1 & 52.8 & 32.6 & 58.2 & 42.4 & 23.0 & 43.4 & 65.6 & 74.4 & 79.9 & 51.1\\ 
% OCNet-(ASP-OC) \citep{OCNet2018yuan} & ResNet50 & 40.2 & 47.3 & 40.2 & 44.4 & 65.0 & 24.1 & 29.9 & 2.71 & 46.3 & 13.6 & 10.3 & 34.6 & 37.9 & 41.4 & 68.1 & 38.0\\ 
% EMANet  & ResNet50 & 55.4 & 63.1 & 68.4 & 66.2 & 82.7 & 56.0 & 18.8 & 42.1 & 58.2 & 41.0 & 33.4 & 38.9 & 46.9 & 46.4 & 78.5 & 47.5\\ 
% CCNet & ResNet50 & 58.3 & 61.4 & 65.7 & 68.9 & 82.9 & 57.1 & 56.8 & 34.0 & 57.6 & 38.3 & 31.6 & 36.5 & 57.2 & 75.0 & 75.8 & 45.9\\ 
% EncodingNet & ResNet50 & 58.9 & 59.7 & 64.9 & 70.0 & 84.2 & 55.2 & 46.3 & 36.8 & 57.2 & 38.7 & 34.8 & 42.4 & 59.8 & 69.8 & 76.1 & 48.0\\ 
UPerNet \citep{UPerNet-2018-jointLearn-multitask} & ResNet50 & 63.8 & 68.7 & 71.0 & 73.1 & 85.5 & 55.3 & 57.3 & 43.0 & 61.3 & 45.6 & 30.3 & 45.7 & 68.7 & 75.1 & 84.3 & 56.2\\ 
HRNet \citep{hrnet2019} & HRNetW18 & 61.5 & 65.9 & 68.9 & 74.0 & 86.9 & 59.4 & 61.5 & 33.8 & 62.1 & 46.9 & 14.9 & 44.2 & 52.9 & \textbf{75.6} & 81.7 & 52.2\\ 
GSCNN \citep{gatescnn-2019iccv} & ResNet50 & 63.4 & 65.9 & 71.2 & 72.6 & 85.5 & 56.1 & 58.4 & 40.7 & 63.8 & 51.1 & 33.8 & 48.8 & 58.5 & 72.5 & 83.6 & 54.4\\
SFNet \citep{SFNet-eccv-2020} & ResNet50 & 64.3 & 68.8 & 71.3 & 72.1 & 85.6 & 58.8 & 60.9 & 43.1 & 62.9 & 47.7 & 30.4 & 47.8 & 69.8 & 75.1 & 83.1 & 57.3\\ 
RANet \citep{RANet-cvpr2019} & ResNet50 & 62.1 & 67.1 & 61.3 & 72.5 & 85.1 & 53.2 & 47.1 & \textbf{45.3} & 60.1 & 49.3 & 38.1 & 41.8 & 70.5 & 58.8 & 83.1 & 55.6\\ 
PointRend \citep{Pointrend-2020cvpr} & ResNet50 & 62.8 & 64.4 &  69.9 &  73.7 &  82.9 &  55.4 &  \textbf{61.1} &  38.5 &  62.3 & 48.1 &  29.8 &  45.0 &  66.0 &  72.7 &  80.7 &  54.0\\ 
FarSeg \citep{FarSeg2020-cvpr} & ResNet50 & 63.7 & 65.4 & 61.8 & 77.7 & 86.4 & 62.1 & 56.7 & 36.7 & 60.6 & 46.3 & 35.8 & 51.2 & 71.4 & 72.5 & 82.0 & 53.9\\ 
PointFlow \citep{Pointflow2021-cvpr} & ResNet50 & 66.9 & 70.3 & 74.7 & 77.8 & 87.7 & 62.2 & 59.5 & 45.2 & 64.6 & 50.2 & 37.9 & 50.1 & 71.7 & 75.4 & 85.0 & 59.3\\
\hline
AF$_2$-AFPN & ResNet50 & \textbf{67.8}  & \textbf{69.5} & \textbf{73.7} & \textbf{80.9} & \textbf{89.9} & \textbf{66.2} & 56.5 & 41.1 & \textbf{67.3} & \textbf{52.0} & \textbf{38.9} & \textbf{53.1} & \textbf{77.9} & 74.4 & \textbf{84.5} & \textbf{59.6}\\
    \bottomrule
  \end{tabular}}
 
\end{table}

\begin{figure}[h]
  \centering
  \includegraphics[width=0.48\textwidth]{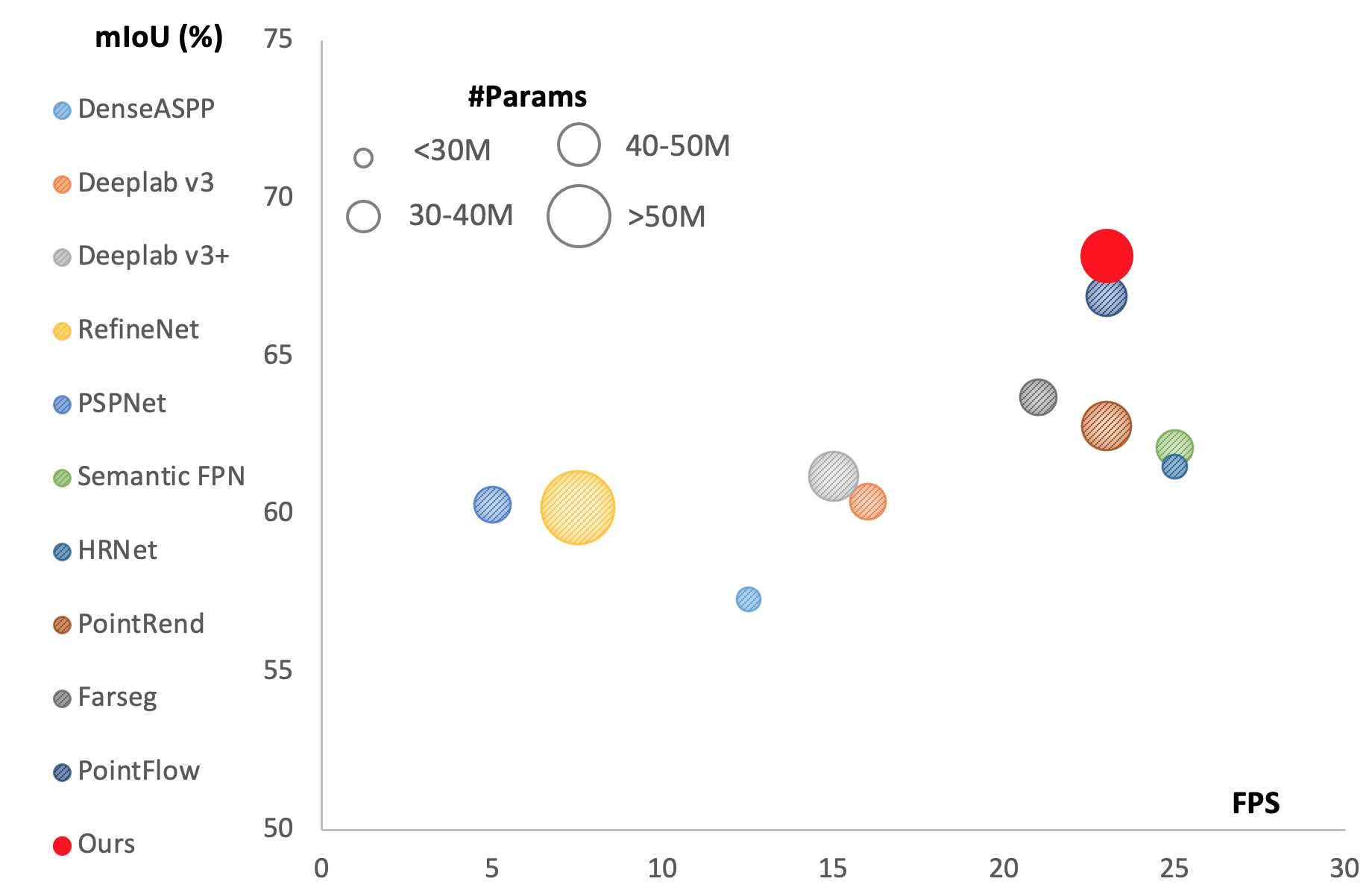}
   \caption{Speed (FPS) versus accuracy (mIoU) on iSAID $val$ set.}
  \label{fig:model efficiency (Speed FPS)}
\end{figure}

\textbf{Efficiency Comparison:} we compare the model size and inference speed on validation set in Figure~\ref{fig:model efficiency (Speed FPS)}. It shows that AF$_2$-AFPN has exceeded most mainstream models in terms of running efficiency.

% \begin{wrapfigure}{r}[0cm]{0.5\textwidth}
% \end{wrapfigure}

\subsection{Results on Vaihingen Dataset}

Vaihingen contains 33 images (of different sizes) and 6 categories have been defined. Following previous work~\citep{Pointflow2021-cvpr}, we adopt large patches as the iSAID dataset.
%  and use more validation images for testing
We utilize 16 images for training and  the rest 17 images for testing.
For training set, the image IDs are 
1, 3, 5, 7, 11, 13, 15, 17, 21, 23, 26, 28, 30, 32, 34, 37.
For validation set, the images IDs are
2, 4, 6, 8, 10, 12, 14, 16, 20, 22, 24, 27, 29, 31, 33, 35, 38.
We crop the images into 768$\times$768 with a sliding window striding 512 pixels, and all the experiments are trained with 200 epochs. Like previous work, we use the mIoU and m-$F_1$ (i.e. the harmonic mean of precision and recall) as the main metric. 
% Following \citep{Pointflow2021-cvpr}, 
% 2 10, 2 11, 2 12, 3 10, 3 11, 3 12, 
% 4 10, 4 11, 4 12, 5 10, 5 11, 5 12, 6 7, 6 8, 6 9, 6 10, 
% 6 11, 6 12, 7 7, 7 8, 7 9, 7 10, 7 11, 7 12.

% 2 13, 2 14, 3 13, 3 14, 4 13, 4 14, 4 15, 
% 5 13, 5 14, 5 15, 6 13, 6 14, 6 15, 7 13.

\begin{figure}[]
  \centering
  \includegraphics[width=1\linewidth]{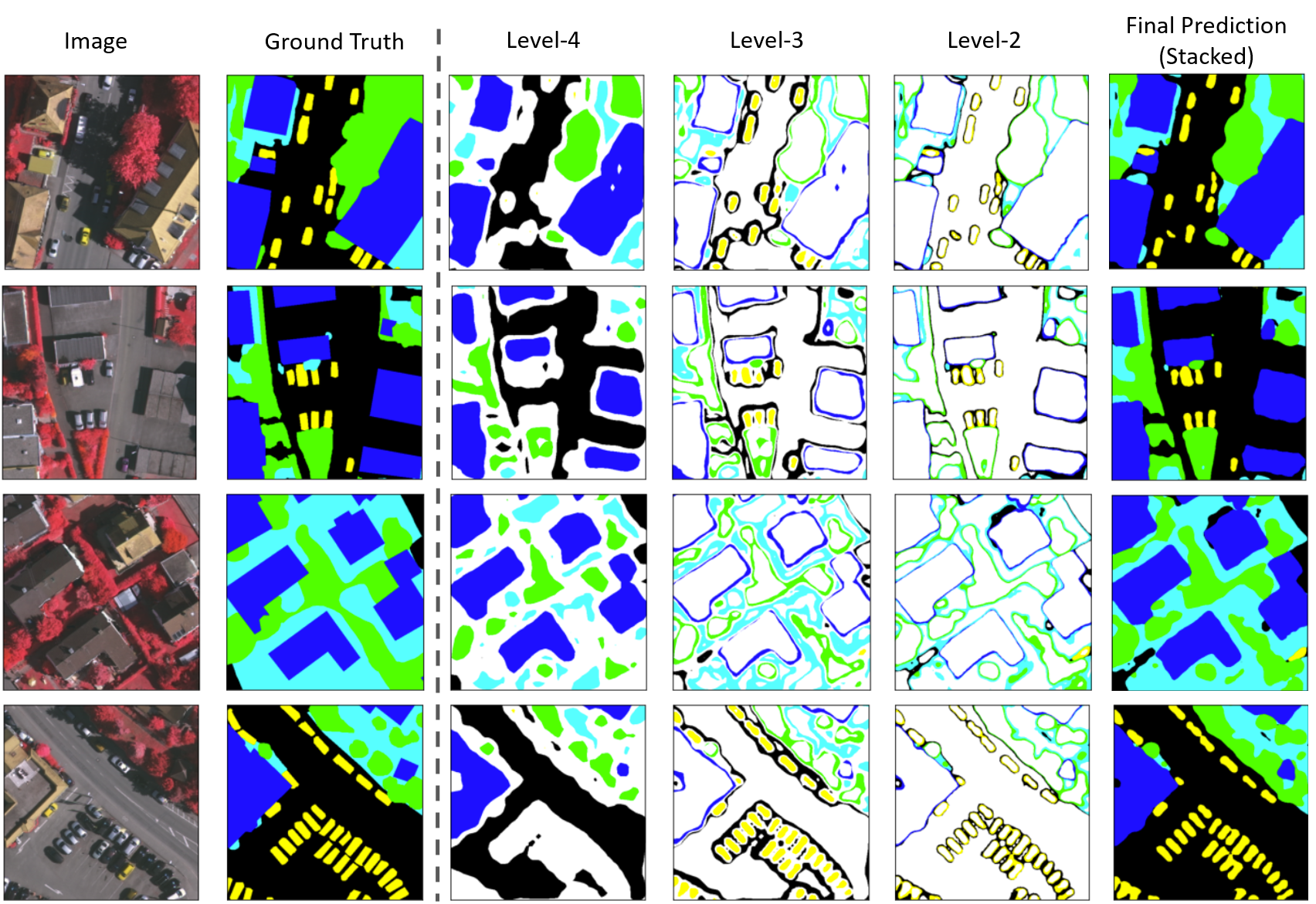}
  \caption{Image Prediction Process Show Case on Vaihingen $val$ set.}
  \label{fig:Image Prediction Process Show Case on vaihingen set}
\end{figure}

The detailed results on Vaihingen Dataset are shown in Tab. \ref{tab:Experimental results on the Vaihingen Dataset}. The process of prediction is shown in Fig. \ref{fig:Image Prediction Process Show Case on vaihingen set}. In Vaihingen dataset, the imbalance between foreground and background is relatively slight. Among all categories, only the cars are tiny, while the background is complex. Our framework also has a big improvement in these categories, which demonstrates the flexibility of AF$_2$.

\begin{table}[]
\renewcommand\arraystretch{1.1}         %表格内部 2 倍行距离
\renewcommand\tabcolsep{3.0pt}
    \centering
    \caption{Experimental results on the Vaihingen Dataset. The results are reported with single scale input. The bold values in each column represent the best entries. The category are defined as: impervious surfaces (Imp.surf.), buildings (Build), low vegetation (Low veg), trees (Tree), cars (Car), cluster/background (Cluster). All the models are trained under the same setting following the PointFlow \citep{Pointflow2021-cvpr}.}
    \label{tab:Experimental results on the Vaihingen Dataset}
    \scalebox{0.8}{
    \begin{tabular}{c|c|c|ccccccc}
    \toprule
\multirow{2}{*}{Method} & \multirow{2}{*}{mIoU(\%)} & \multirow{2}{*}{mean-$F_1$} & \multicolumn{6}{c}{$F_1$ per category} \\
& & & Imp.surf. & Build. & Low veg. & Tree & Car & Cluster\\ 
\hline
FCN  \citep{FCN2015-cvpr} & 64.2 & 75.9 & 87.6 & 91.6 & 77.8 & 84.6 & 73.5 & 40.3\\ 
PSPNet \citep{pspnet(PPM)2017-cvpr}  & 65.1 & 76.8 & 88.4 & 92.8 & 79.2 & 85.9 & 73.5 & 41.0\\ 
OCNet(ASP-OC) \citep{OCNet2018yuan} & 65.7 & 77.4 & 88.8 & 92.9 & 79.2 & 85.8 & 73.9 & 43.8\\ 
Denseaspp \citep{denseaspp-2018cvpr} & 64.7 & 76.4 & 87.3 & 91.1 & 76.2 & 83.4 & 77.1 & 43.3\\ 
Deeplabv3+ \citep{deeplabv3p-2018-eccv} & 64.3 & 76.0 & 88.7 & 92.8 & 78.9 & 85.6 & 72.4 & 37.6\\ 
% DANet  & 65.3 & 77.1 & 88.5 & 92.7 & 78.8 & 85.7 & 73.7 & 43.2\\ 
% CCNet  & 64.3 & 75.9 & 88.3 & 92.5 & 78.8 & 85.7 & 73.9 & 36.3\\ 
SemanticFPN \citep{PanopticFPN-2019cvpr}  & 66.3 & 77.6 & 89.6 & 93.6 & 79.7 & 86.3 & 75.7 & 40.7\\ 
UPerNet \citep{UPerNet-2018-jointLearn-multitask}  & 66.9 & 78.7 & 89.2 & 93.0 & 79.4 & 86.0 & 74.9 & 49.7\\ 
HRNet-W18 \citep{hrnet2019} & 66.9 & 78.2 & 89.2 & 92.6 & 78.7 & 85.7 & 77.1 & 45.9\\ 
GSCNN \citep{gatescnn-2019iccv} & 67.7 & 79.5 & 89.4 & 92.6 & 78.8 & 85.4 & 77.9 & 52.9\\ 
SFNet \citep{SFNet-eccv-2020} & 67.6 & 78.6 & 90.0 & \textbf{94.0} & 80.3 & \textbf{86.5} & 78.9 & 41.9\\ 
% EMANet  & 65.6 & 77.7 & 88.2 & 92.7 & 78.0 & 85.7 & 72.7 & 48.9\\ 
RANet \citep{RANet-cvpr2019} & 66.1 & 78.2 & 88.0 & 92.3 & 79.1 & 86.0 & 78.8 & 53.1\\ 
% EncodingNet  & 65.5 & 77.4 & 88.6 & 92.5 & 78.5 & 85.7 & 73.6 & 45.5\\ 
PointRend \citep{Pointrend-2020cvpr} & 65.9 & 78.1 & 88.2 & 92.4 & 78.9 & 84.5 & 73.5 & 51.1\\ 
FarSeg \citep{FarSeg2020-cvpr} & 65.7 & 78.0 & 88.0 & 92.0 & 78.2 & 85.2 & 73.3 & 51.5 \\
PointFlow \citep{Pointflow2021-cvpr} & 70.4 & 81.9 & \textbf{90.1} & 93.6 & 77.7 & 85.4 & \textbf{80.0} & \textbf{64.6}\\ 
\hline
AF$_2$-AFPN &\textbf{70.5} & \textbf{82.1} & 89.6 & 93.4 & \textbf{79.8} & 86.2 & 79.8 & 63.5 \\
\bottomrule
    \end{tabular}}
\end{table}

\subsection{Results on Potsdam Dataset}

Potsdam contains 38 images (of different sizes) and 6 categories have been defined. 
Following~\citep{Pointflow2021-cvpr}, we utilize 24 images for training, and the image IDs are 
2\_10, 2\_11, 2\_12, 3\_10, 3\_11, 3\_12, 4\_10, 4\_11, 4\_12, 5\_10, 5\_11, 5\_12, 6\_7, 6\_8, 6\_9, 6\_10, 6\_11, 6\_12, 7\_7, 7\_8, 7\_9, 7\_10, 7\_11, 7\_12.
We utilize the another 14 images for test, and the image IDs are 
2\_13, 2\_14, 3\_13, 3\_14, 4\_13, 4\_14, 4\_15, 5\_13, 5\_14, 5\_15, 6\_13, 6\_14, 6\_15, 7\_13.
We crop the images into 896$\times$896 with a sliding window striding 512 pixels, and all the experiments are trained with 80 epochs. Like previous work, we use the mIoU and m-$F_1$ as the main metric.

\begin{figure}[]
  \centering
  \includegraphics[width=1\linewidth]{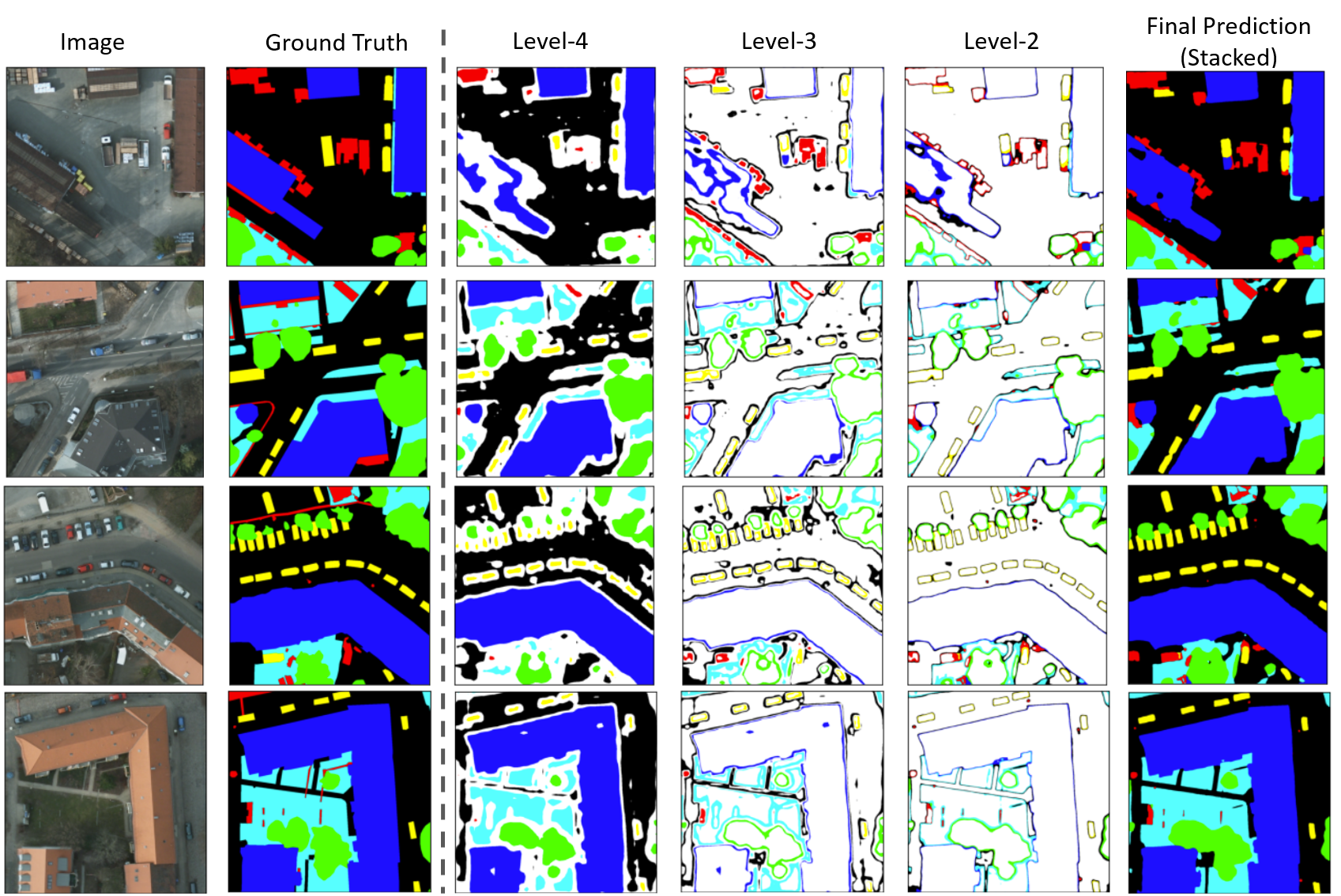}
  \caption{Image Prediction Process Show Case on Potsdam $val$ set.}
  \label{fig:Image Prediction Process Show Case on potsdam set}
\end{figure}

We show detailed results in Tab. \ref{tab:Experimental results on the Potsdam Dataset} and the process of prediction in Fig. \ref{fig:Image Prediction Process Show Case on potsdam set} for Potsdam dataset.

\begin{table}[]
\renewcommand\arraystretch{1.1}         %表格内部 2 倍行距离
\renewcommand\tabcolsep{3.0pt}
    \centering
    \caption{Experimental results on the Potsdam Dataset. The results are reported with single scale input. The bold values in each column represent the best entries. The category are defined as: impervious surfaces (Imp.surf.), buildings (Build), low vegetation (Low veg), trees (Tree), cars (Car), cluster/background (Cluster). All the models are trained under the same setting following the PointFlow \citep{Pointflow2021-cvpr}.}
    \label{tab:Experimental results on the Potsdam Dataset}
    \scalebox{0.8}{
    \begin{tabular}{c|c|c|ccccccc}
    \toprule
\multirow{2}{*}{Method} & \multirow{2}{*}{mIoU(\%)} & \multirow{2}{*}{mean-$F_1$} & \multicolumn{6}{c}{$F_1$ per category} \\
& & & Imp.surf. & Build. & Low veg. & Tree & Car & Cluster\\ 
\hline
FCN \citep{FCN2015-cvpr} & 73.1 & 83.1 & 90.2 & 94.7 & 84.1 & 85.6 & 89.2 & 54.8\\ 
PSPNet \citep{denseaspp-2018cvpr} & 73.9 & 83.9 & 90.8 & 95.4 & 84.5 & 86.1 & 88.6 & 58.0\\ 
OCnet(ASP-OC) \citep{OCNet2018yuan} & 74.2 & 84.1 & 90.9 & 95.5 & 84.8 & 86.0 & 89.2 & 58.2\\ 
Denseaspp \citep{denseaspp-2018cvpr} & 73.9 & 83.9 & 90.8 & 95.4 & 84.6 & 86.0 & 88.5 & 58.1\\ 
Deeplabv3+ \citep{deeplabv3p-2018-eccv}& 74.1 & 83.9 & 91.0 & 95.6 & 84.6 & 86.0 & 90.0 & 56.2\\ 
% DAnet & 74.0 & 83.9 & 91.0 & 95.6 & 84.9 & 86.2 & 88.7 & 57.0\\ 
% CCnet & 73.8 & 83.8 & 90.7 & 95.5 & 84.7 & 86.0 & 88.5 & 57.3\\ 
SemanticFPN \citep{PanopticFPN-2019cvpr} & 74.3 & 84.0 & 91.0 & 95.5 & 84.9 & 85.9 & 90.4 & 56.3\\ 
UPerNet \citep{UPerNet-2018-jointLearn-multitask} & 74.3 & 84.0 & 90.9 & 95.7 & 85.0 & 86.0 & 90.2 & 56.2\\ 
HRNet-W18 \citep{hrnet2019} & 73.4 & 83.4 & 90.4 & 94.9 & 84.2 & 85.4 & 90.0 & 55.5\\ 
GSCNN \citep{gatescnn-2019iccv} & 73.4 & 84.1 & 91.4 & 95.5 & 84.8 & 85.8 & 91.2 & 55.9\\ 
SFNet \citep{SFNet-eccv-2020} & 74.3 & 84.0 & 91.0 & 95.5 & 85.1 & 86.0 & 90.9 & 55.5\\ 
% EMANet & 72.9 & 83.1 & 90.4 & 94.9 & 84.2 & 85.7 & 88.3 & 55.1\\ 
RANet \citep{RANet-cvpr2019} & 73.8 & 83.9 & 90.8 & 92.1 & 84.3 & 86.8 & 88.9 & 56.0\\ 
% EncodingNet & 73.4 & 83.5 & 90.6 & 95.1 & 84.5 & 86.0 & 88.2 & 56.6\\ 
PointRend \citep{Pointrend-2020cvpr}& 72.0 & 82.7 & 89.8 & 94.6 & 82.8 & 85.2 & 85.2 & 58.6\\ 
FarSeg \citep{FarSeg2020-cvpr} & 73.4 & 83.3 & 90.7 & 95.2 & 84.3 & 85.3 & 90.2 & 54.1\\
PointFlow \citep{Pointflow2021-cvpr} & 75.4 & 84.8 & 91.5 & 95.9 & 85.4 & 86.3 & 91.1 & 58.6 \\
\hline
% AF$_2$-AFPN & 74.6 & 84.2 & 91.1 & 95.5 & 84.9 & 85.9 & 91.7 & 56.0 \\
AF$_2$-AFPN & 74.9 & 84.4 & 91.1 & 95.7 & 85.0 & 86.2 & 91.5 & 57.0 \\
\bottomrule
    \end{tabular}}
\end{table}

\end{document}